**Enhancing Readmission Prediction with Deep Learning: Extracting Biomedical Concepts from Clinical Texts**

Rasoul Samani[1,†], Mohammad Dehghani[2,*,†], *Fahime Shahrokh*[1]

[1]*School of Electrical and Computer Engineering, Isfahan University of Technology, Isfahan, Iran*
[2]*School of Electrical and Computer Engineering, University of Tehran, Tehran, Iran*

* Correspondence: dehghani.mohammad@ut.ac.ir

† These authors contributed equally to this work.

## Abstract

Hospital readmission, defined as patients being re-hospitalized shortly after discharge, is a critical concern as it impacts patient outcomes and healthcare costs. Identifying patients at risk of readmission allows for timely interventions, reducing re-hospitalization rates and overall treatment costs. This study focuses on predicting patient readmission within less than 30 days using text mining techniques applied to discharge report texts from electronic health records (EHR). Various machine learning and deep learning methods were employed to develop a classification model for this purpose. A novel aspect of this research involves leveraging the Bio-Discharge Summary Bert (BDSS) model along with principal component analysis (PCA) feature extraction to preprocess data for deep learning model input. Our analysis of the MIMIC-III dataset indicates that our approach, which combines the BDSS model with a multilayer perceptron (MLP), outperforms state-of-the-art methods. This model achieved a recall of 94% and an area under the curve (AUC) of 75%, showcasing its effectiveness in predicting patient readmissions. This study contributes to the advancement of predictive modeling in healthcare by integrating text mining techniques with deep learning algorithms to improve patient outcomes and optimize resource allocation.

## 1. Introduction

The healthcare sector of the country is among the fields that impose substantial costs on the government and insurance organizations annually [1]. The digitization of patients' medical records, in addition to enhancing the quality of medical services provided to citizens, will establish the foundation for numerous cost savings in the country's healthcare expenditure [2, 3]. Online healthcare services rely significantly on EHR to store, exchange, and manage patients' medical information effectively [4]. One of the examined metrics in the realm of improving treatment quality and achieving financial savings is the utilization of electronic patient files to monitor the readmission rate of patients in hospitals [5].

Hospital readmission refers to the process in which a patient, having been discharged from the hospital, returns and is readmitted within a specific and relatively short period [6]. The readmission rate of a hospital is currently regarded as an indicator of the hospital's performance quality in certain circumstances. A high level of this metric has adverse effects on patient care costs [7]. Consequently, medical service centers have introduced a plan to mitigate hospital readmissions [8]. The goal of this approach is to enhance patient care quality and reduce medical care expenses. In instances where the rate of patient readmissions exceeds acceptable levels, hospitals may face financial penalties [9, 10].

Different studies have investigated the influential factors contributing to the rate of patient readmissions to hospitals and have aimed to predict the probability of a patient being readmitted within a short period after discharge (e.g., 30 days, 70 days, and 90 days) [11-13]. Most predictive approaches have primarily considered the patient's most recent hospital visit. One of the effective methods for preventing rapid and repeated patient readmissions to the hospital is the utilization of artificial intelligence (AI) and its predictive approaches. These approaches enable the prediction and identification of patients who are highly likely to return to the hospital. Subsequently, more accurate and appropriate treatment decisions can be made to reduce this probability [14].

Text mining, as a subset of artificial intelligence, aims to tackle specific tasks like identifying relevant documents and extracting important details from them, contributing to problem-solving efforts within various domains [15]. In healthcare, many studies have utilized text mining methods to either classify patients into pre-defined groups or extract valuable insights from various sources such as clinical notes [16], academic publications [17], and social media where patients share their experiences and information [18].

In this study, utilizing clinical notes, a range of machine learning and deep learning models including logistic regression, random forest, K-nearest neighbors (KNN), support vector machine (SVM), Gaussian naive Bayes, and MLP were employed to construct a predictive model for patient readmission. Our study makes several notable contributions:

(1) **Wide Range of Models:** In contrast to previous approaches primarily relying on machine learning, our research systematically compared a wide range of machine learning and deep learning models. As a result, we identified the most accurate predictive model.
(2) **Advanced Text Representation:** In addition to Term Frequency-Inverse Document Frequency (TF-IDF), we employed the BDSS model for word embedding to encode clinical text data. Furthermore, we utilized PCA for dimensionality reduction, enhancing computational efficiency.
(3) **Size of Dataset:** Diverging from prior approaches, our study boldly deployed the proposed models across the entire dataset, even in the face of significant imbalance.
(4) **Addressing Imbalanced Datasets:** In stark contrast to prior methodologies that leaned on undersampling or oversampling, our study confidently refrained from employing data balancing techniques. Remarkably, our models exhibited commendable performance, underscoring their robustness and efficacy.
(5) **Input Features:** Our investigation exclusively utilized textual features, intentionally avoiding any dependence on demographic attributes. This deliberate choice allowed us to focus on the intrinsic content of the data, unencumbered by external factors.

The rest of paper is organized as follows: Section 2 provides a review of related work. Section 3 outlines the proposed model, which includes introducing the dataset, detailing preprocessing steps, and describing the learning models employed. Section 4 presents the results obtained from the experiments conducted and provides a discussion on the findings. Finally, Section 5 concludes the paper.

## 2. Related Works

In recent years, different studies have focused on machine/deep learning applications in various aspects of the healthcare system [19, 20]. In light of the significance attributed to this issue, several studies have been conducted to investigate the factors influencing the rate of hospital readmissions. Andriotti et al. [21] has examined the relationship between the duration of a patient's hospital stay and the likelihood of their re-hospitalization. The findings reveal that if the patient's hospital stay is excessively short, the probability of their subsequent return to the hospital increases.

Mardini and Raś [22] highlight hospital readmissions as a key factor contributing to the rise in healthcare expenditures by the US government. This research employed a clustering procedure to group patients based on the similarity of their medical diagnoses. Subsequently, an algorithm was introduced, which provided treatment recommendations to physicians, resulting in a reduction in the number of patients being readmitted to the hospital. These studies shed light on the importance of understanding the relationship between hospital stay duration, medical clustering, and readmission rates, contributing to the development of strategies for mitigating readmissions and optimizing healthcare outcomes.

Brindise and Steele [23] developed a model using machine learning techniques applied to patient admission data and clinical texts. This model successfully predicted the likelihood of patient readmission prior to their discharge. The approach employed in this particular study involved employing a Bayesian network for patient classification which achieved an AUC index of 71%.

In Ben-Assuli et al. study [24], various methods, including linear regression, decision tree, and neural networks, were applied to the patient's electronic file data and the frequency of visits to the emergency department. Through the comparison of these methods, the enhanced decision tree approach demonstrated the highest result, achieving 85% accuracy compared to the other models.

In Zheng et al. research [25], the patient's clinical data, such as age, length of hospitalization, gender, and insurance coverage, were utilized as numerical or categorical features. A classification model called PSO-SVM was designed. This model achieved the highest accuracy of 78% compared to linear regression and random forest approaches.

Furthermore, Hammoudeh et al. [26] developed a deep learning algorithm to predict readmission among diabetic patients within three time periods: less than 30 days, between 30 and 70 days, and more than 70 days. The algorithm utilized medical records of diabetic patients spanning a ten-year period. With an 8% improvement in prediction scores compared to other methods, the model achieved an AUC of 79%.

Assaf and Jayousi [27] explored the prediction of patient readmission within 30 days, using various patient demographic data, such as gender, age, admission status, as well as laboratory results and disease diagnosis codes. The study achieved an AUC of 66%.

In Moerschbacher and He study [28], patient EHR data, including demographic information, laboratory results, and discharge reports, were processed. The feature engineering process of patient discharge texts employed the bag of word approach. This study achieved a performance of 75.7% in the AUC.

## 3. Methodology

An overview of the proposed method can be observed in Figure 1. Following the preparation and pre-processing of the required data, including the conversion of textual data into numerical vectors, machine learning and deep learning models were designed and implemented to predict readmission.

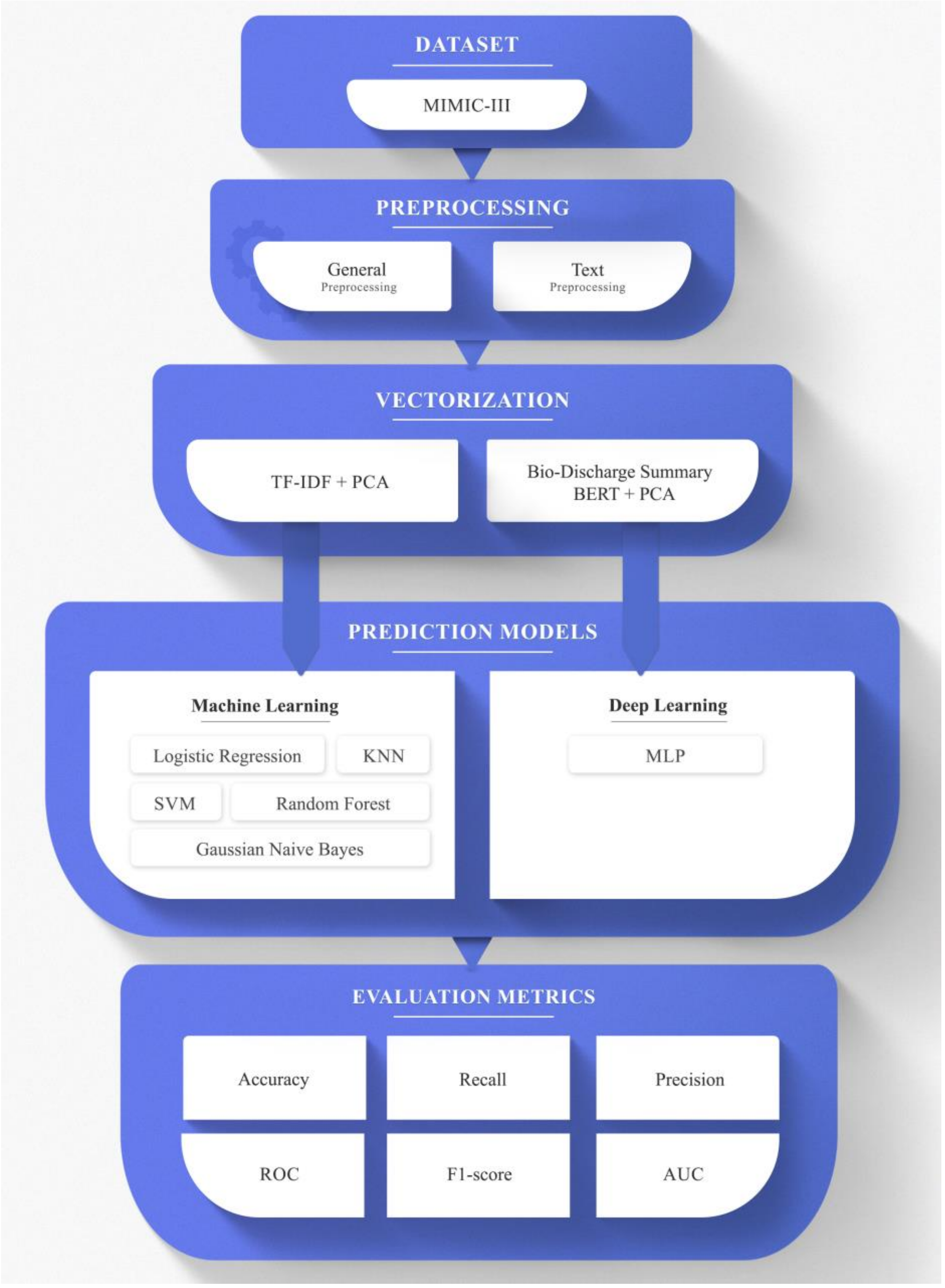

Figure 1: Overview of the proposed method.

### 3.1 Data

For this research, the MIMIC-III dataset (Medical Information Mart for Intensive Care III) [29] was utilized. This freely available dataset encompasses data from over 50,000 patients (with their identifying information removed) admitted to a US hospital between 2001 and 2012. In this study, two key tables from this dataset were employed, namely the admission table and the clinical report table. The admission table contains information regarding patient admissions, while the clinical report table includes written records such as medical history and discharge instructions documented by attending physicians in the patient's file.

### 3.2 Data preprocessing

Data preprocessing is a crucial step in training a model, as it directly impacts its performance. The quantity, quality, and diversity of data influence the effectiveness of machine and deep learning algorithms. To improve the reliability of the algorithms, it's vital to carefully select and preprocess the target data from the original dataset [30]. Efficient data preprocessing techniques such as dimensionality reduction and data transformation play a critical role in enhancing the extraction of valuable insights and knowledge from datasets [31]. We conducted two preprocessing phases on the input data. Initially, general preprocessing techniques were applied to the dataset records and columns, with the objective of improving data quality and structure. Following this, text preprocessing methods were utilized.

### 3.2.1. General preprocessing

As depicted in Figure 2, admission table and clinical report table were preprocessed separately to ensure that each dataset underwent tailored preprocessing steps suited to its specific characteristics and requirements.

For admission table, following preprocessing steps were employed:

- Data related to patients whose status was recorded as birth or death were removed.
- Additional columns were added to the admission table to record the date and type of the patient's next admission to the hospital. These columns were populated for each patient's admission, with the information of their subsequent admission arranged in chronological order. For the last patient admission, the values of these columns were predicted using the research approach. Subsequently, only the rows of data were retained where the patient's admission was elective, excluding cases where the patient was admitted as an emergency.
- A new column was added to store the number of days until the next admission, which serves as the target variable for this approach. The objective is to predict samples for which the value of this parameter will be less than 30 days, indicating readmission within a short time frame.

The preprocessing steps applied to the clinical report table were as follows:

- All notes for “Discharge summary” were filtered.
- Null texts were removed.

To further refine the dataset and reduce missing values in the considered columns, the admission tables and discharge summary reports were merged based on the patient's admission ID.

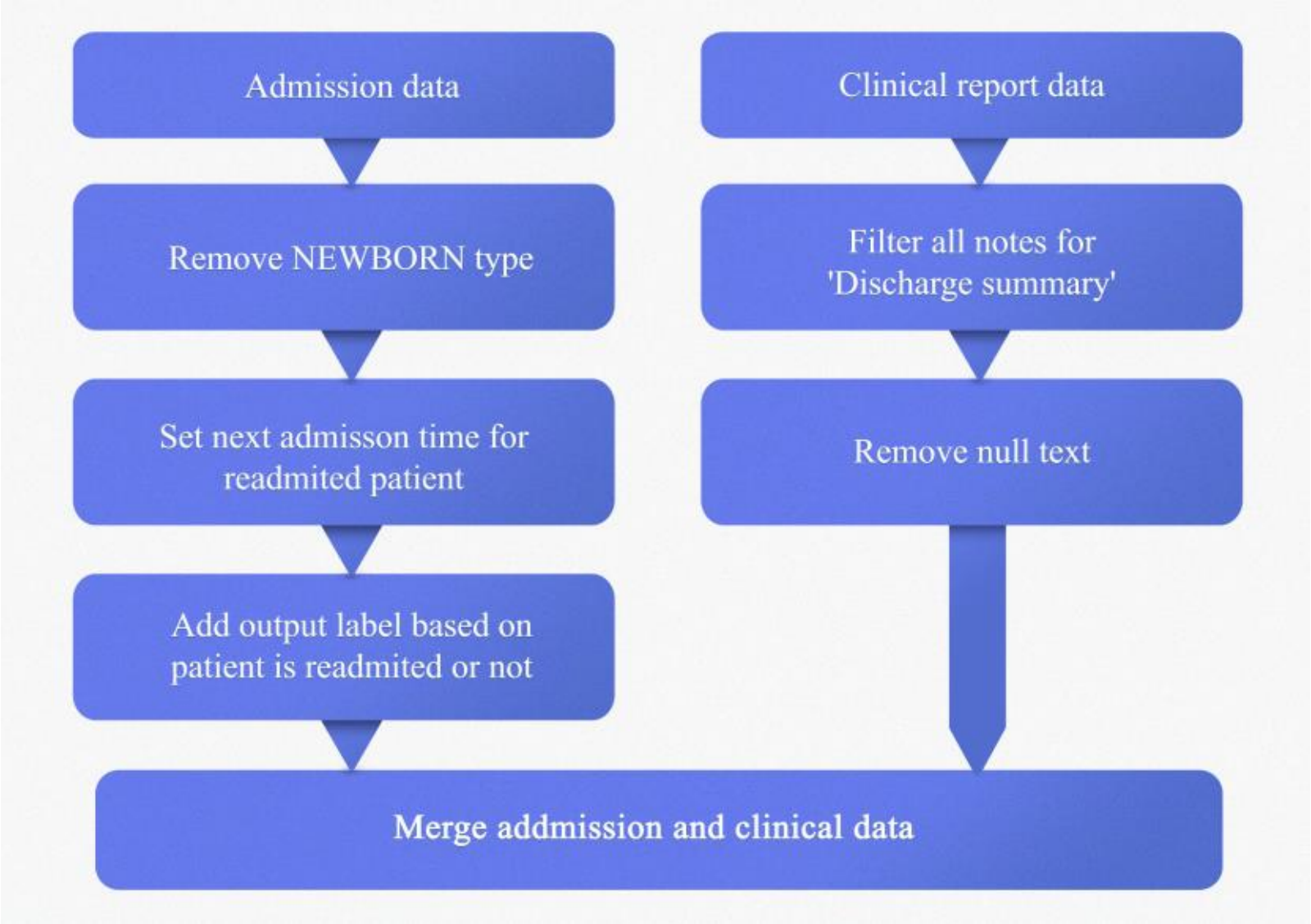

Figure 2: General preprocessing.

### 3.2.2. Text preprocessing

Text preprocessing involves cleaning, transforming, and preparing textual data to make it suitable for analysis and modeling [32]. In Figure 3, the text preprocessing steps are illustrated.

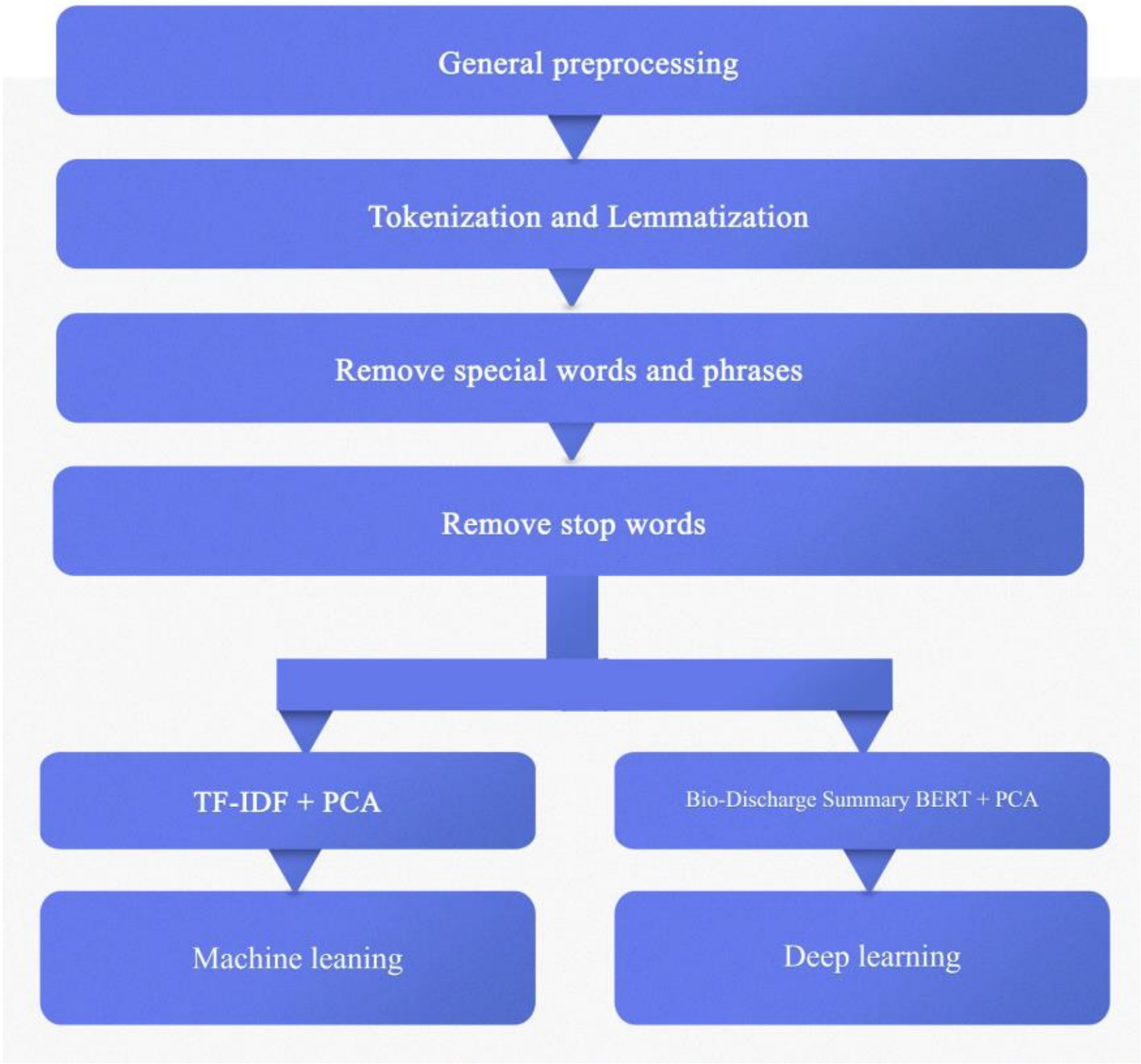

Figure 3: Text preprocessing.

Stages of text preprocessing encompass a range of crucial tasks, including:

- **Tokenization:** Tokenization is the process of breaking down text into individual components such as characters, words, phrases, or other elements known as tokens [33].
- **Lemmatization:** Groups inflected forms of words together based on their lemma.
- **Stop word removal:** Stop words, such as "a", "the", "in", "with", which are often common terms in a language and lack meaningful information [34], are excluded.
- **Special words removal:** Commonly occurring words and phrases, including titles, unidentified brackets, and drug dosages, are eliminated from all notes during preprocessing.

In Figure 4 (a), the word cloud of the original data reveals a predominant focus on drug-related terms, dosage information, and titles, which are frequently repeated in the text but may not contribute significantly to predicting readmission. However, certain specific words, such as laboratory results, hold more relevance for conducting analysis. Therefore, we conducted preprocessing steps to filter out less relevant terms and retain more valuable ones. The results of this preprocessing are depicted in Figure 4 (b), where drug-related terms have been removed, and more informative words that are important for both doctors and our predictive model have been retained. This selective filtering enhances the quality of the data input for our analysis, enabling more accurate predictions of readmission classification.

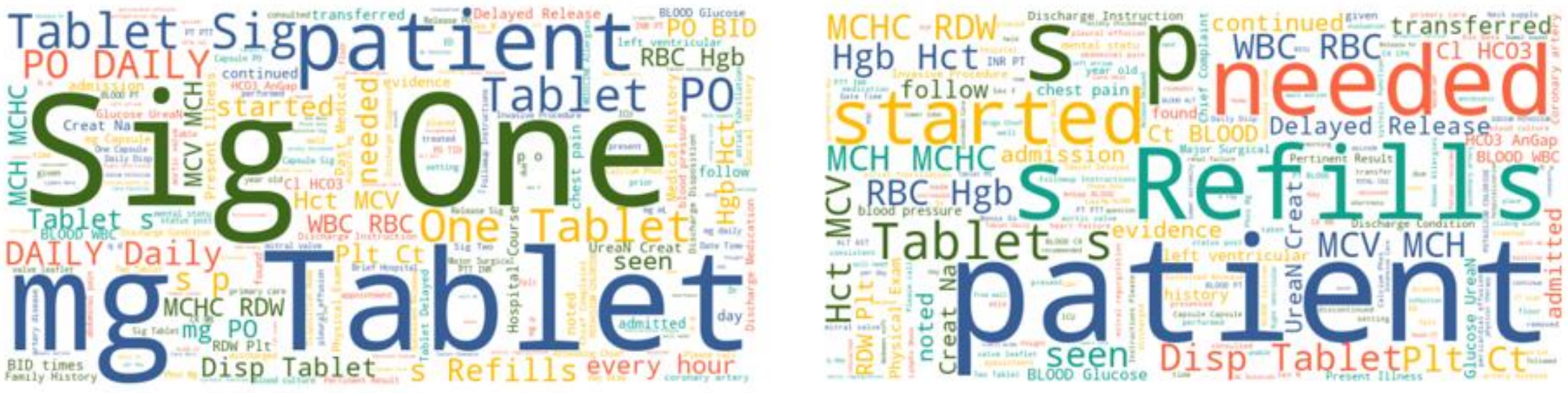

(a) (b)

Figure 4: Removal of irrelevant words.

As machines can't process free text directly, numerical conversion is essential for analysis and modeling. Raw text data must undergo transformation into numerical formats before being utilized for training model [30]. Accordingly, two methods were employed for text vectorization in the subsequent preprocessing phase. TF-IDF was utilized for machine learning models, while BDSS model was applied for deep learning model.

### *TF-IDF*

TF-IDF, is a widely used method for weighting terms in text data. It involves assigning numerical scores to words based on their frequency within a document and their infrequency across the entire corpus, indicating their importance in the context of the document and the corpus as a whole [35]. In this study, we set the max-df parameter to 0.8 and the min-df parameter to 5. Max-df denotes the threshold for the maximum document frequency of terms, used for removing terms that appear too frequently. Conversely, min-df indicates the minimum document frequency required for terms to be considered, with terms appearing in fewer documents than the specified min-df typically excluded from analysis.

The vocabulary size of 35097 is exceptionally high, posing challenges in terms of computational resources and processing time required for thorough analysis. To address this issue, dimensionality reduction techniques are indispensable. One such technique is PCA, which effectively reduces the dimensionality of the data while preserving essential information. In our approach, we applied PCA with a consideration of 50 principal components.

### *Discharge Summary BERT*

Trained on clinical text derived from roughly 2 million notes within the MIMIC-III v1.4 database, BDSS is designed to focus solely on discharge summaries, ensuring the corpus is customized to align with downstream tasks [36]. We utilized the BDSS model [37], and trained exclusively on discharge summaries extracted from the MIMIC database. This approach guarantees that the model is finely tuned for tasks related to clinical text analysis.

After completing preprocessing steps, the dataset's maximum text length is 7479 tokens, with a mean length of 1317 words, and the length of the third quarter is 1735 words. Selecting an appropriate input size for vectorization is crucial. If the input size is too high, it can lead to increased computational complexity, while too low of a size may result in the loss of valuable information. Given that BDSS's input size is 512, we choose an input size of 2048, which is four times the size of BDSS's input and can accommodate the third quarter of the text data. To ensure consistency in input size, we pad data with smaller sizes and truncate data with longer sizes, thus maintaining all inputs at the same size (2048) while preserving important information.

Following the initial data input, which has a size of 2048, we divide it into four segments, each of size 512. These segments serve as input to the BDSS model for vectorization. The output from BDSS has dimensions of 512 x 768 for each input. To obtain a concise representation, we compute the average for each entry, resulting in a 768-dimensional vector. The final output comprises four such vectors, each sized at 768, yielding a total size of 3072. Subsequently reached 50 features with the help of PCA. Finally, we fed these reduced features into a MLP model. This approach allowed us to effectively capture essential information while minimizing computational complexity.

### 3.3. Predictive models

In this study, multiple data mining algorithms and deep learning were employed to create prediction model.

**Logistic regression:** Logistic regression, a statistical technique, finds extensive use in binary classification tasks, particularly in health sciences studies where the focus lies on disease states (diseased or healthy) and decision-making scenarios (yes or no) [38].

**Random forest:** Random forest utilizes a structure composed of numerous decision trees, with predictions from each tree combined to forecast the value of a variable [39].

**KNN:** KNN is a model that leverages the values of the nearest samples in the training data to determine the category or value of a given sample [40].

**SVM:** Using SVM, data are transformed into a high-dimensional feature space where separating hyperplanes are constructed to maximize the margin between data points and the hyperplane, effectively delineating them into distinct classes [41]. This process facilitates robust classification by ensuring clear boundaries between different classes in the feature space, enhancing the model's ability to generalize to unseen data.

**Gaussian Naive Bayes:** Naive Bayes is a probabilistic classifier that employs Bayes' theorem to estimate the probability of a given set of features belonging to a specific label. It calculates the conditional probability of event A occurring given the individual probabilities of A and B, as well as the conditional probability of event B. This approach assumes that features are independent [42]. Gaussian Naive Bayes is a variant of the Naive Bayes classifier that assumes features to follow a Gaussian distribution.

**MLP:** The MLP adopts a feed-forward structure, comprising an input layer, one or more hidden layers, and an output layer interconnected with neurons. Each neuron in one layer is linked to every neuron in the next layer, facilitating the transmission of input signals from inputs to outputs through the hidden neurons. The number of neurons in the input layer corresponds to the number of input variables in the dataset after preprocessing, while the number of neurons in the output layer matches the number of classes in the dataset (which is two in our study). These networks typically include hidden layers with nonlinear transfer functions, enabling them to learn both linear and nonlinear relationships between input and output vectors [43, 44].

By employing these diverse algorithms, this study aims to explore their effectiveness in predicting patient readmission rates and identifying the most suitable approach for the given dataset and research objectives.

## 4. Evaluation

### 4.1. Evaluation metrics

In binary classification tasks, data instances are typically classified as either positive or negative. A positive label signifies the presence of readmission, while a negative instance indicates no-readmission. Each binary label prediction can be categorized into one of four possibilities: a true positive (TP) occurs when a positive outcome is correctly predicted, a true negative (TN) happens when a negative outcome is correctly predicted, a false positive (FP) arises when a negative instance is wrongly predicted as positive, and a false negative (FN) occurs when a positive instance is incorrectly predicted as negative [45].

The primary evaluation metrics for binary classification are accuracy, precision, recall, and F1-score. Accuracy represents the percentage of correctly classified instances among all instances (Equation 1). Precision measures the proportion of instances classified as positive among all instances predicted as positive (Equation 2). Recall, also known as sensitivity, assesses the ability of the model to identify all truly positive instances (Equation 3). Finally, the F1-score is a harmonic mean of precision and recall, providing a balanced assessment of the model's performance (Equation 4) [46].

$$Accuracy = \frac{TP + TN}{TP + FP + TN + FN} \tag{1}$$

$$Precission = \frac{TP}{TP + FP} \tag{2}$$

$$Recall = \frac{TP}{TP + FN} \tag{3}$$

$$F1_score = \frac{2 * Precission * Recall}{Precission + Recall} \tag{4}$$

Two additional valuable metrics include ROC (Receiver Operating Characteristic) and AUC. The ROC curve is constructed by plotting the true positive rate against the false positive rate. This curve consistently increases within the unit square, bounded by the points (0, 0) and (1, 1) [47]. In addition to the ROC curve, the area under it (AUC) serves as another valuable evaluation metric. This metric spans from 0 to 1, providing insight into the overall performance of the classification model [48].

### 4.2. Results and discussion

The dataset initially comprised 51,113 records, which underwent preprocessing resulting in 49,083 records. All these records were utilized to construct classification models. It's noteworthy that the dataset exhibits a high degree of imbalance, and to maintain realism and promote better generalization, no balancing techniques were applied. The data was then partitioned into three subsets: 70% for training, amounting to 34,358 records, 15% for validation, containing 7,363 records, and the remaining 15% for testing, also consisting of 7,362 records. Table 1 provide the distribution of each class.

Table 1: statics of dataset.

| | **Readmission** | **No-readmission** |
|---|---|---|
| All data after preprocessing | 2942 | 46141 |
| Train | 2059 | 32299 |
| Validation | 442 | 6921 |
| Test | 441 | 6921 |

Table 2 presents the results obtained from various classifiers employed in our study. Notably, the Final Method, which combines the BDSS model with MLP, outperformed the state-of-the-art models in terms of AUC. Furthermore, this model, along with logistic regression, achieved the highest accuracy, recall, and F1-score, underscoring the continued relevance of machine learning models. In Figure 5, the ROC curve illustrates the performance of different models, with the Final Method achieving an impressive AUC of 75%, surpassing all other models. Remarkably, logistic regression exhibited superior performance with a rate of 73.2%, outperforming alternative machine learning techniques.

In the medical domain, metrics like recall and AUC play a crucial role in evaluating AI models. Recall, which measures the ability of a model to correctly identify positive cases, is particularly important in healthcare settings where identifying all potential cases is paramount. Similarly, AUC provides an overall measure of model performance and is widely used for assessing predictive models in medical applications. The Final Method, leveraging the BDSS model, is considered the best model due to its superior performance in terms of recall and AUC. This model is trained on discharge summaries data and harnesses the power of BDSS, which is pre-trained on a large corpus of text data and is adept at understanding the semantic nuances of text.

Table 2: Results of proposed models.

| **Model** | **Accuracy** | **Precision** | **Recall** | **F1-score** | **AUC** |
|---|---|---|---|---|---|
| KNN | 0.638 | 0.906 | 0.638 | 0.733 | 0.623 |
| SVM | 0.861 | 0.898 | 0.861 | 0.878 | 0.649 |
| Gaussian Naive Bayes | 0.649 | 0.91 | 0.649 | 0.741 | 0.66 |
| Random Forest | **0.94** | 0.883 | **0.94** | 0.909 | 0.728 |
| Logistic Regression | **0.94** | **0.944** | **0.94** | **0.911** | 0.732 |
| **Final Method (BDSS+MLP)** | **0.94** | 0.884 | **0.94** | **0.911** | **0.75** |

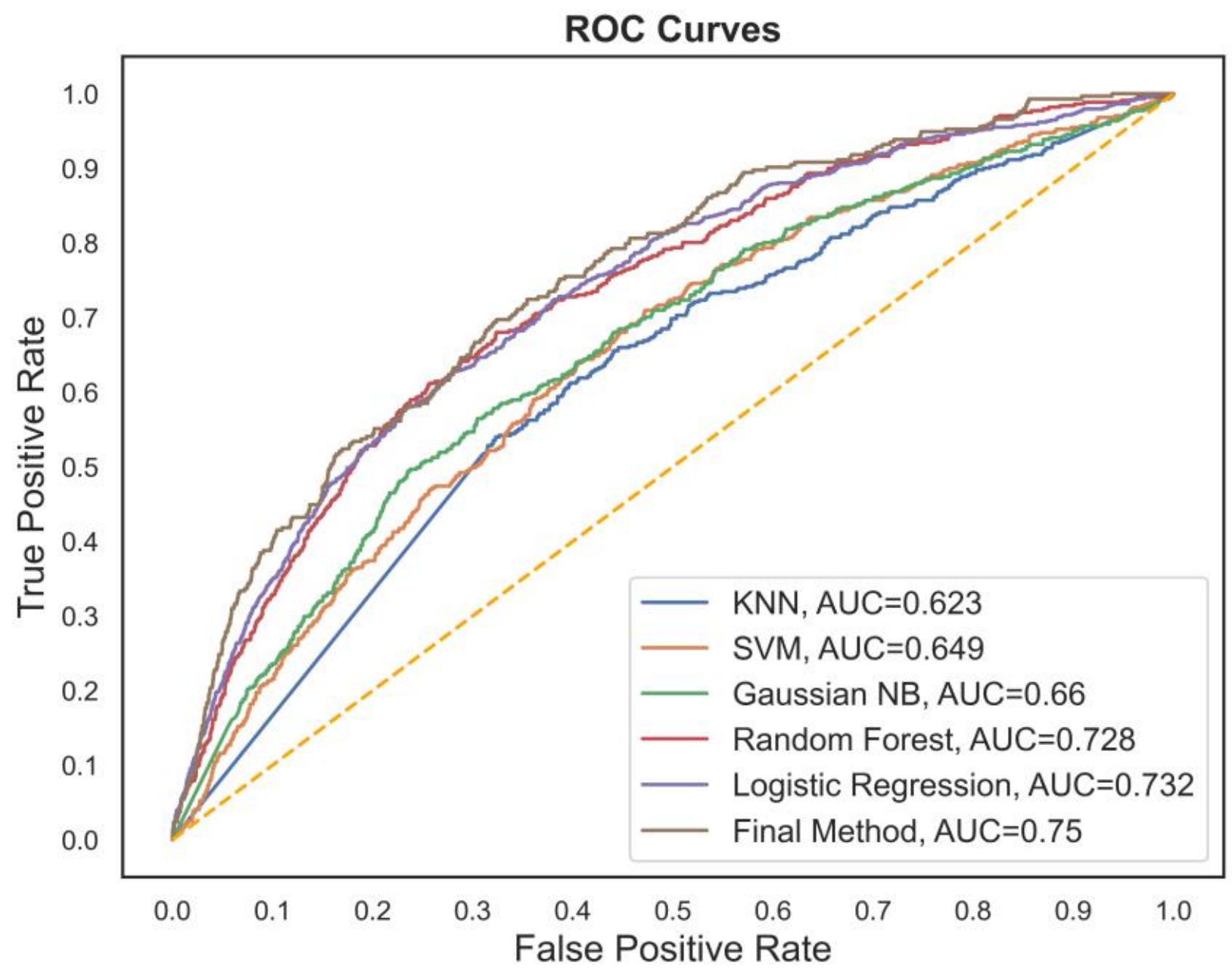

Figure 5: ROC curve.

One advantage of the logistic regression model is its clarity and interpretability. To gain insights into the model's decision-making process, we extracted and presented the features that exerted the most significant impact on the outcomes, as shown in Figure 6. As observed, words such as "milliliter," "mg," and "chronic" had the greatest influence on categorizing patients as readmitted. This can be attributed to the prescription of various drugs with specific doses by the medical practitioners during the patient's discharge. The higher the number of prescribed drugs, the higher the likelihood of patient readmission. Conversely, the presence of words like "without," "family," "negative," "normal," and "transferred" in the patient's discharge text had the most substantial impact on categorizing patients as non-returning to the hospital.

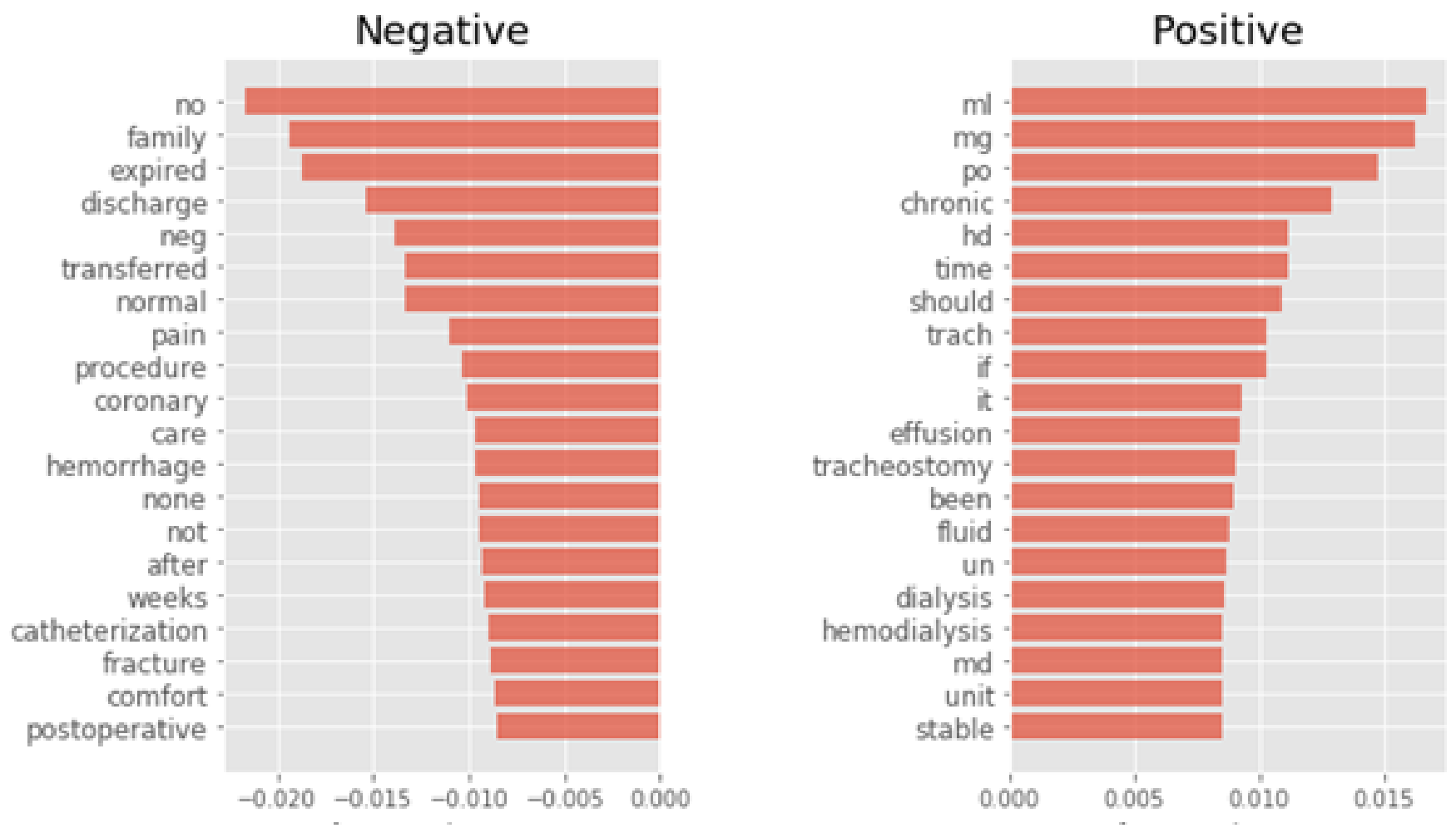

Figure 6: The most effective words in the classification of patients in the logistic regression model.

Several previous studies have investigated models for predicting ICU readmission, with logistic regression consistently demonstrating favorable results, achieving AUC rates of 65% [49], 66% [50], and 70% [51]. However, a recent study by Orangi-Fard et al. [41], utilized various machine learning techniques on the MIMIC-III dataset to predict patient readmission. Their SVM-RBF model achieved an AUC rate of 74%. It's worth noting that Orangi-Fard et al. only utilized a portion of the dataset (4000 for training and 6000 for validation), balanced data, and employed 825 features. In contrast to previous approaches, our study took a comprehensive approach by utilizing the entire dataset, including imbalanced data. Additionally, we focused solely on textual features, omitting other factors such as demographics. This deliberate choice allowed us to gain deeper insights into the specific aspects we aimed to explore. Furthermore, while previous studies solely relied on machine learning models, our study also incorporated deep learning methods. This highlights the novelty and potential advantages of leveraging deep learning techniques in predicting ICU readmission. Table 3 provide a comparison with existing methods based on AUC metric.

Table 3: Comparing with previous studies.

| **Ref.** | **AUC** |
|---|---|
| [49] | 65% |
| [50] | 66% |
| [51] | 70% |
| [41] | 74% |
| Our method | 75% |

## 5. Conclusion

Medical data, particularly EHR data, presents a rich source for text mining studies. These studies hold promise in various healthcare applications. Reducing ICU readmission rates is paramount for hospitals to enhance patient outcomes, conserve ICU resources, and curtail healthcare expenses. In this study, we aimed to leverage patient discharge reports, which offer detailed insights into a patient's medical history, current condition, and treatment recommendations, to develop a predictive model for ICU readmission. Our proposed deep learning-based model demonstrated superior performance compared to traditional machine learning models, achieving higher AUC. For future research, exploring alternative deep learning architectures beyond MLP could be beneficial. Additionally, Large Language Models (LLM) can be considered for creating predictive models and conducting comparative analyses with deep learning models. To enhance their effectiveness, we recommend considering the use of larger input data and leveraging advanced models like the LongFormer. Additionally, incorporating summarization techniques during the pre-processing stage can further improve the quality of input data.